\documentclass[letterpaper, 10 pt, conference]{ieeeconf}  

\IEEEoverridecommandlockouts 

\usepackage{graphicx}
\usepackage{glossaries}
\usepackage{amsfonts}
\usepackage{amsmath}
\usepackage{amssymb}
\usepackage{caption}
\usepackage{tabularx}
\usepackage{algorithm2e}
\usepackage{textcomp}
\usepackage{tikz}   
\usepackage{graphicx}
\usepackage{subfigure}
\usepackage{siunitx}
\usepackage{subcaption}
\usepackage{booktabs} 
\usepackage{algpseudocode}
\RestyleAlgo{ruled}

\usepackage{amsmath}

\newacronym{ica}{ICA}{Internal Carotid Artery}
\newacronym{mca}{MCA}{Middle Cerbral Artery}
\newacronym{pca}{PCA}{Posterior Cerbral Artery}
\newacronym{pcom}{PComm}{Posterior Communicating Artery}
\newacronym{rms}{RMS}{Root Mean Square}

\usepackage{xcolor}
\usepackage{etoolbox}
\usepackage{tikz}
\usepackage[normalem]{ulem}

\newtoggle{colorMode}

\newcommand{\myBlue}[1]{%
    \iftoggle{colorMode}{\textcolor{blue}{#1}}{#1}%
}
\newcommand{\myRed}[1]{%
  \iftoggle{colorMode}{\textcolor{red}{\sout{#1}}}{\unskip\ignorespaces}%
}

\newcommand{\myReview}[1]{%
    \iftoggle{colorMode}{%
        \colorbox{yellow}{\textbf{\textcolor{black}{#1}}}%
    }{}%
}

\togglefalse{colorMode}

\begin{document}

\title{\LARGE \bf An Anatomy-specific Guidewire Shaping Robot for Improved Vascular Navigation}

\author{Aabha Tamhankar, Jay Patil, Giovanni~Pittiglio
\thanks{FuTURE Lab, Department of Robotics Engineering, Worcester Polytechnic Insitute (WPI), Worcester, MA 01605, USA. Email: {\tt\small \{astamhankar, jpatil1,gpittiglio\}@wpi.edu}
\newline This work was supported by the Worcester Polytechnic Institute (WPI), Department of Robotics Engineering.}%
} 

\maketitle

\begin{abstract}
Neuroendovascular access often relies on passive microwires that are hand-shaped at the back table and then used to track a microcatheter to the target. Neuroendovascular surgeons determine the shape of the wire by examining the patient's pre-operative images and using their experience to identify anatomy-specific shapes of the wire that would facilitate reaching the target. This procedure is particularly complex in convoluted anatomical structures and is heavily dependent on the surgeon's level of expertise. Towards enabling standardized autonomous shaping, we present a bench-top guidewire shaping robot capable of producing navigation-specific desired wire configurations. We present a model that can map the desired wire shape into robot actions, calibrated using experimental data. We show that the robot can produce clinically common tip geometries (C, S, Angled, Hook) and validate them with respect to the model-predicted shapes in 2D. Our model predicts the shape with a \gls{rms} error of 0.56\:mm across all shapes when compared to the experimental results. We also demonstrate 3D tip shaping capabilities and the ability to traverse complex endoluminal navigation from the petrous \gls{ica} to the \gls{pcom}. 
\glsresetall
\end{abstract}


\IEEEpeerreviewmaketitle
\section{Introduction}
\label{sec:introduction}
\begin{figure}[t]
    \centering
    \includegraphics[width=\columnwidth]{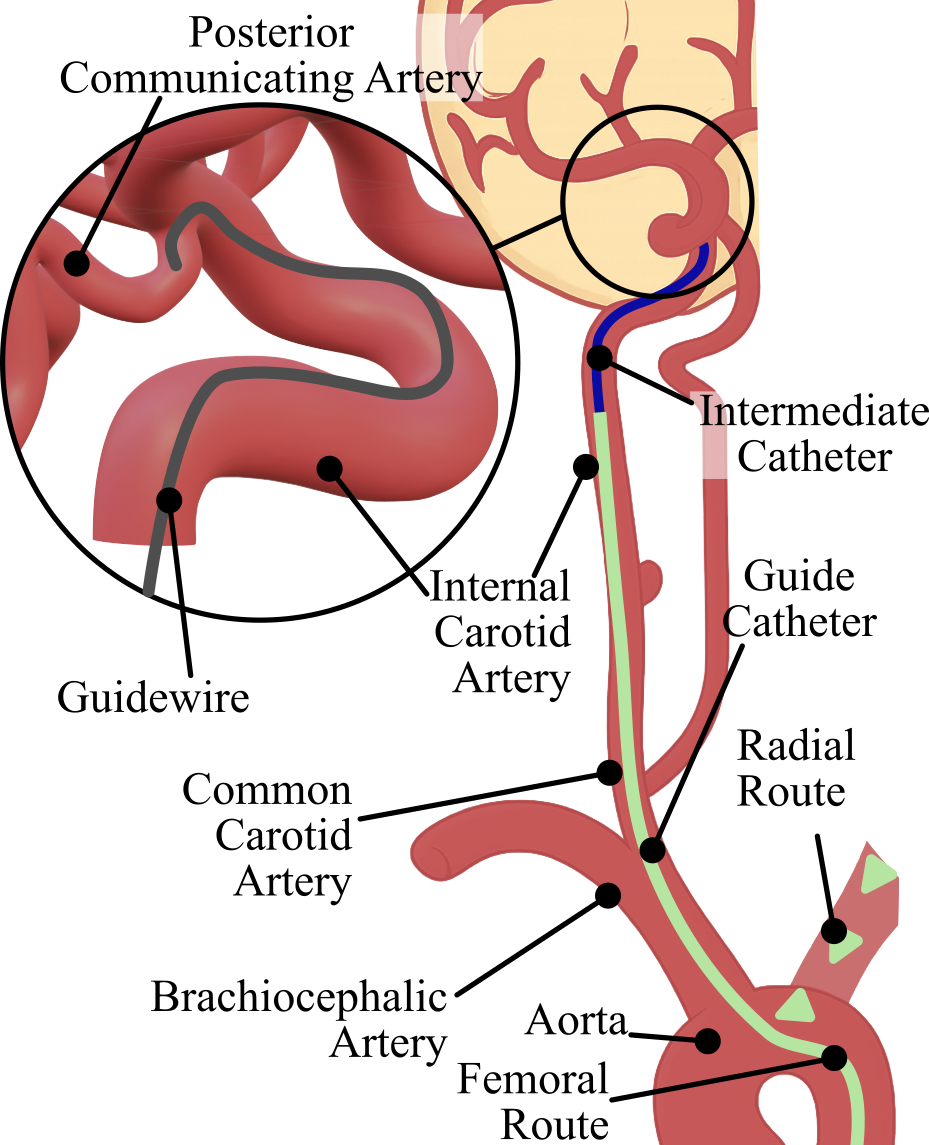}
    \caption{Catheterization for neuroendovascular treatments with alternative access routes via arm (radial) or leg (femoral). Example of guidewire navigation past the \gls{ica} into the \gls{pcom}.}
    \label{fig:main_figure}
\end{figure}

Neuroendovascular procedures, such as mechanical thrombectomy for ischemic stroke, aneurysm embolization, and targeted embolotherapy, are complex interventions that require navigating tortuous vasculature while operating under strict time constraints. These procedures are performed via radial or femoral access after creating an endovascular access path through which tools are deployed (Fig. \ref{fig:main_figure}). 


The clinical standard is to use a set of up to four coaxial tools, from distal to proximal: guide wire, microcatheter, intermediate catheter, and guide catheter \cite{Bhatia2020CurrentStroke, Jang2016MonoplaneProcedures}. These passive tools are advanced through the vasculature and navigated under biplane fluoroscopy. Navigation leverages the natural shape of the tools and intentional contact with the vessel' walls, while accounting for millimeter-scale vessel clearances, severe tortuosity, and dynamic vessel motion, all while managing radiation and contrast exposure.  

The main element that drives navigation to the target location is the guidewire, which is advanced to the target and allows the microcatheter to track along it. The microcatheter is then often used to deliver treatment, e.g., in aneurysm treatments, or to further advance the intermediate catheter for aspiration in mechanical thrombectomy. 

In routine practice, operators select pre‐shaped wire geometries or manually shape wire and microcatheter tips to match an anticipated three-dimensional vascular geometry, often tortuous, branching, and highly variable across patients \myRed{, and manually shape the wire tip with mandrels on the back table \cite{Liu2023NavigatingEmbolization}}. \myBlue{This manual customization is typically performed on the back table, where operators bend the tip over mandrels and apply steam to set the shape \cite{Liu2023NavigatingEmbolization}.
Prior work has explored patient-specific 3D-printed molds to provide a static template for shaping the tools to the vessel anatomy \cite{Xu2020MicrocatheterCoiling}. owever, these static workflows impose significant fabrication delays and lack the versatility to adapt to intra-operative changes, while traditional manual shaping remains inherently subjective and difficult to standardize.
} \myReview{R2-4}
Shaping neuroendovascular guidewires to match patient-specific anatomy to achieve optimal navigation and stability is a complex and highly skill-dependent process. Small deviations in curvature or orientation can result in failure to access the target vessel, excessive contact with the vessel wall, or loss of support for subsequent catheter advancement \cite{Pandhi2021UnderstandingRadiology}. This process typically relies on the surgeon’s experience and visual estimation from preoperative or fluoroscopic images rather than quantitative feedback, which makes reproducibility difficult. Furthermore, wires must maintain their shape once inserted, but they still need to be flexible enough to follow the vessel path, requiring a delicate balance of stiffness and elasticity that is challenging to achieve consistently.

Active distal steering has been demonstrated in research settings using magnetic actuation \cite{Dreyfus2024DexterousAccess,Brockdorff2025HybridApplications,Kim2022TeleroboticManipulation} and tendon/cable‐driven concepts \cite{Lis2022DesignRobot,Abah2024Self-SteeringInterventions}. While these approaches can enhance reachability and targeting, they often demand significant investment in equipment or specialized manufacturing and integration, complicating the integration to neurointerventional workflows. Patient‐specific, piecewise‐magnetized devices have similarly shown promise in other endoluminal domains \cite{Pittiglio2022Patient-SpecificEndoscopy,Pittiglio2023PersonalizedLungs}, but on‐demand fabrication in the angiography suite remains impractical at present.    

To enable seamless clinical translation, we target a robotic platform that works with current practice—shaping standard, commercially available shapeable guidewires, rather than fabricating patient-specific devices. This preserves existing workflows and supply chains, and shifts the burden to accurate modeling of wire–anatomy interactions so that shapes can be planned and executed reliably. In passive systems without tip steering, this requires online localization within preoperative images and predictive models of how wire–vessel contact governs steering \cite{Tamhankar2025TowardsPlanning,Chi2020CollaborativeLearning,Karstensen2023RecurrentArch}.

Building on this premise, we recently proposed a contact-aware planner that makes physics-based predictions from preoperative imaging and known wire properties \cite{Tamhankar2025TowardsPlanning}. This planner has shown to enable repeatability in navigating the aortic arch in-vitro and has the potential to be used to predict the most appropriate shapes for navigating further inside the anatomy. The goal of the present paper is to provide a complementary robotic system capable of shaping currently used guidewires, eventually driven by the knowledge of the patient anatomy and contact-aware wire navigation.

In this work, we design an autonomous wire-shaping robot that generates clinically relevant geometries by coordinating three primitives: axial insertion, axial rotation, and programmable segment-wise bending via a custom gripper. The robot executes a roll–bend–advance cycle to synthesize 3D tip shapes from low-dimensional commands (bend per segment and plane clocking), mirroring back-table technique. The shaping sequence is executed autonomously based on the desired shape parameters

We present a model to map a desired wire geometry into the corresponding robot actuation commands. The mapping is derived from the kinematics of wire bending and refined using experimental calibration data to compensate for material nonlinearity and fixture compliance. The model allows for the prediction of the resulting wire shape from a given set of actuator inputs and can be inverted to compute the required actuation to achieve a specified target shape.

Model accuracy was validated experimentally using common clinical wire shapes, including C, S, angled, and hook tips. Predicted and measured shapes were compared in 2D, showing an average root-mean-square deviation of \SI{0.56}{\milli\meter} across all configurations. Additional tests demonstrated the system’s ability to generate 3D tip geometries and to reproduce shapes suitable for navigation from the petrous \gls{ica} to the \gls{pcom}, confirming the model’s generalizability across clinically relevant designs.

\myBlue{We present the development of a novel robotic platform that enables autonomous, repeatable shaping of standard off-the-shelf guidewires, eliminating the need for patient-specific device fabrication. Additionally, we contribute a data-driven kinematic model that compensates for material nonlinearity to predict wire geometry from robot actions. Finally, we demonstrate the clinical utility of the system through the experimental validation of complex 3D helical tip shaping for navigation in realistic neurovascular anatomies.}



\section{Guidewire Shaping Robot Design}
\label{sec:shaping_mechanism}

\begin{figure}
    \centering
    \includegraphics[width=\columnwidth]{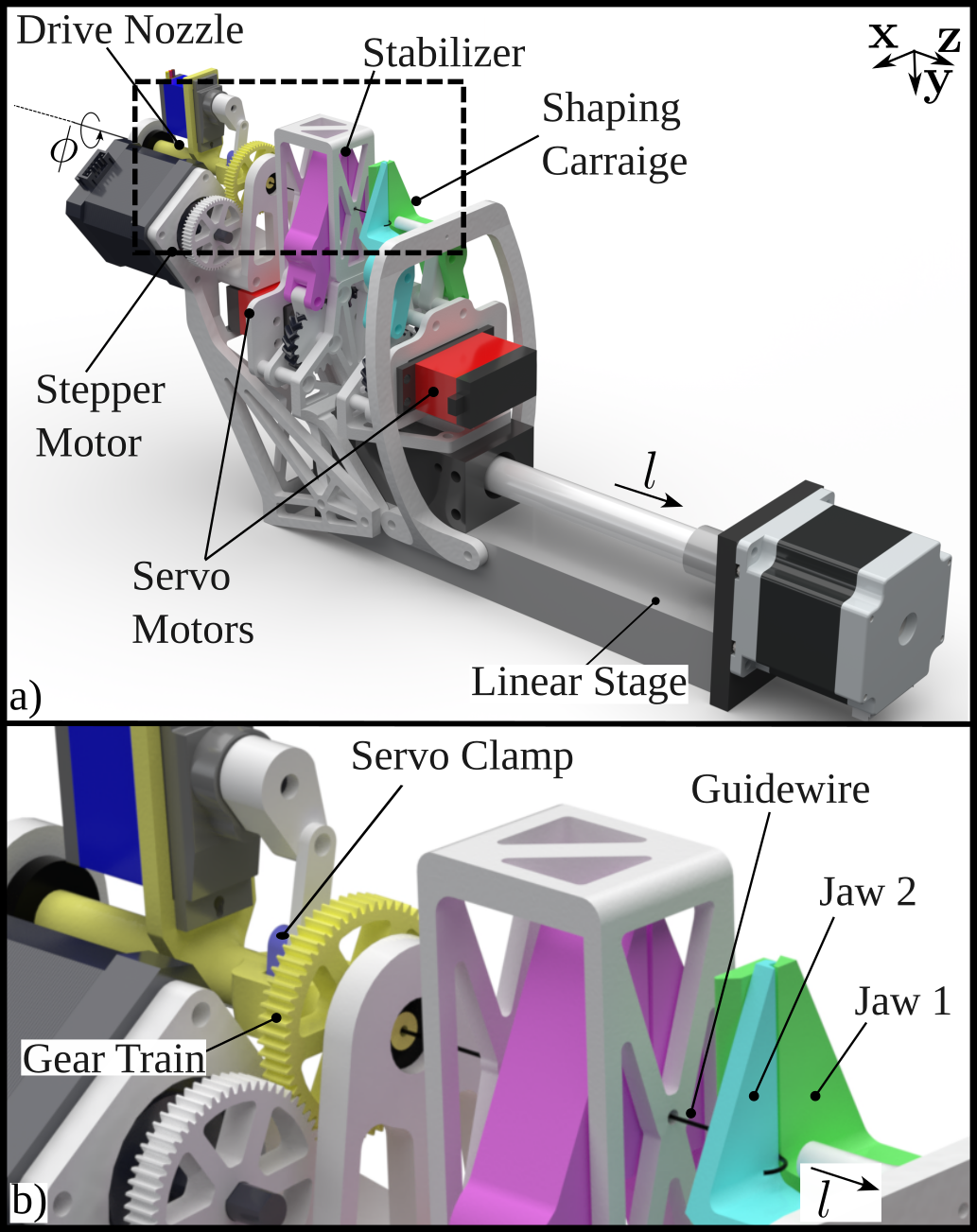}
    \caption{The proposed guidewire shaping robot.}
    \label{fig:mechanism}
\end{figure}
The proposed guidewire shaping robot automates distal guidewire shaping by reproducing the two actions clinicians perform by hand—planar bending of the shapeable tip to set curvature and axial rotation of the wire to choose the bending plane. Figure~\ref{fig:mechanism} labels the three modules of the mechanism arranged in line with the wire: a proximal \emph{Drive Nozzle} that axially rotates and feeds the wire, a \emph{Stabilizer} that holds the wire shaft steady during bending, and a distal \emph{Shaping Carriage} that applies a controlled bend over a short segment of wire.

The \emph{Drive Nozzle} is a short, stiff tube that is concentric with the wire. A stepper motor rotates the tube via a gear train, while a miniature servo clamp inside gently grips the wire so that the commanded rotation of the tube is transmitted as a pure axial roll of the wire.

Immediately downstream, the \emph{Stabilizer} is a passive gripper with a compliant insert. It closes during bending to prevent slips and off-axis deflections, and opens during rotation and feed. Its role is to establish a clean mechanical boundary: the wire is supported but not crushed, so the bending moment applied by the \emph{Shaping Carriage} is concentrated in the distal shape-able zone.

The \emph{Shaping Carriage} is mounted on a precision linear stage that is parallel to the wire. Its two jaws are designed such that Jaw 1 presents a broad, nearly planar surface that supports the wire, while the opposing Jaw 2 presents a narrow line of contact. When the carriage closes and advances a short distance, the opposed contacts “pinch” the wire and create a localized bending moment that plastically deforms the intended segment of the tip. Two inputs govern the local bend: (i) the jaw approach/dwell, which we refer to as $\beta_k$, which sets the bend angle \(\theta_k\), and (ii) the carriage stroke, which indexes the wire forward of the amount $\delta_k$. During shaping, the \emph{Stabilizer} grips the shaft to prevent slipping; during feeding, the \emph{Stabilizer} releases, and the closed jaws temporarily hold the wire while the \emph{Shaping Carriage} translates by the required length. The carriage then reopens and returns to its home position to begin shaping the next segment. 

The shaping operation proceeds as a $k$-th roll-bend-advance cycle for $n$ times, $k=1,2,\dots,n$. Starting from a straight wire, the controller: (i) opens the \emph{Stabilizer} and \emph{Shaping Carriage}; (ii) rolls the wire to the desired bending plane by commanding the \emph{Drive Nozzle} to angle \(\phi_k\); (iii) closes the \emph{Stabilizer}; (iv) closes the \emph{Shaping Carriage}, controlling the servo angle $\beta_k$, and advances along its linear guide to impose the programmed bend $\theta_k$ on the current segment; (v) retracts the \emph{Shaping Carriage} and opens the \emph{Stabilizer}; and (vi) \emph{closes the Shaping Carriage} and feeds the wire of $\delta_k$ by the required segment length to expose the next shaping zone. 

Together, these steps are inspired by the surgeon’s “pinch-and-pull” method, which localizes the bend and then indexes the guidewire tip to the next segment. Repeating this cycle generates a sequence $\left\{\phi_k, \beta_k, \delta_k\right\}_{k=1}^n$ that produces planar or multi-planar geometries. A real-time demonstration is provided in the attached supplementary video, where the robot autonomously shares planar C and S tips using the $\left\{\phi_k, \beta_k, \delta_k\right\}_{k=1}^n$ sequence.

For the construction of the robot, we use off-the-shelf drive elements with known performance. The shaping carriage runs on a CBX-1605 precision linear stage driven by a NEMA-23 stepper, with a positioning accuracy of \(\pm\SI{0.003}{\milli\meter}\) along the wire axis. Proximal axial rotation is supplied by a NEMA-17 stepper (\SI{1.8}{\degree} per full step) coupled through a 60-tooth, module-0.5 spur-gear train to the \emph{drive Nozzle}. A compact MG92B micro-servo actuates the nozzle clamp, so commanded tube rotation transfers as pure wire roll. The gripper jaws on the \emph{Shaping Carriage} are each actuated by DS3218 servos and linked via custom herringbone gears.

Because rotation and bending are physically decoupled, the robot yields consistent inter-segment bend angles and axial rotation, allowing for direct comparison between commanded sequences and realized shapes, as detailed in the next section.

\section{Wire Shaping Model}
\label{sec:wire_model}
\begin{figure}
    \centering
    \includegraphics[width=0.75\columnwidth]{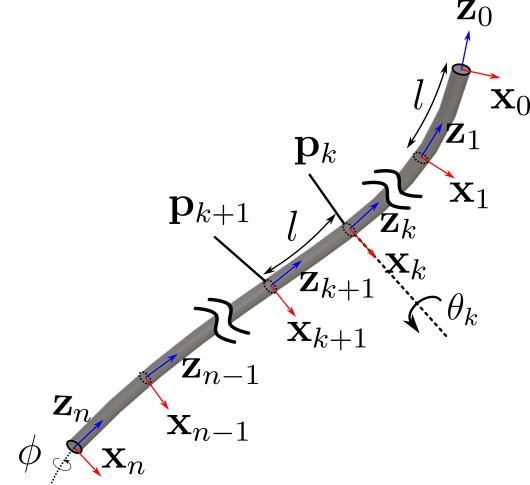}
    \caption{Schematic representation of the shapable wire.}
    \label{fig:wire}
\end{figure}

We derive a guidewire shaping model that maps the robot's actions $\left\{\phi_k, \beta_k, \delta_k\right\}_{k=1}^n$ which enable forming a desired shape of the wire. We describe the shape through the wire's centerline $\mathbf{p}(s): s\rightarrow\mathbb{R}^3, \: s\in [0, L]$, with $s$ length parameter and $L$ total length of the wire. This shown in Fig. \ref{fig:wire}.

We discretize the centerline into $n$ segments of arch-length $l$, such that $L = nl$. The $k$-th segment is described using the constant curvature approximation, where the $z$ axis is assumed to be tangent to the centerline and the pinching action $\beta_k$ occurs along the $y$ axis. We assume that perfect and accurate pinching consequently shapes the wire to bend purely around the $x$ axis; thus, in the reference frame $k+1$, the centerline of the $k$-th segment can be described as  

\begin{equation}
    ^{k+1}\mathbf{p}_k -\:^{k+1}\mathbf{p}_{k+1} = l\text{exp}(\theta_k[\mathbf{e}_1]^\wedge)\mathbf{e}_3
\end{equation}
where $\theta_k$ is the amount by which the segment bends around $\mathbf{e}_1$; $\text{exp}(\cdot)$ is the exponential map of $SO(3)$; $[\mathbf{v}]^\wedge = \left(\mathbf{v}\times\mathbf{e}_1 | \mathbf{v}\times\mathbf{e}_2 | \mathbf{v} \times \mathbf{e}_3\right)$; and $\times$ indicate the cross product. The bending amount $\theta_k$ depends on the mechanical properties of the wire, its mechanical parameters, material composition, and the strength of pinching arising from the action $\beta_k$.

Since the shaping occurs from the most distal segment -- segment 1 in Fig. \ref{fig:wire} -- and the \emph{Stabilizer} (Fig. \ref{fig:mechanism}) prevents side motion of the wire, we can assume that, during the shaping of segment $k$, all the segments before it are straight; hence, $^n\mathbf{p}_{k+1}=l(n-k)\mathbf{e}_3$, in the base frame, implying 

\begin{eqnarray}
 ^{k+1}\mathbf{p}_k &=& \left(^{k+1}\mathbf{R}_n\:^n\mathbf{p}_{k+1} + l\text{exp}\left(\theta_k[\mathbf{e}_1]^\wedge\right)\right)\mathbf{e}_3 \\
     &=& l\left((n-k) \:^{k+1}\mathbf{R}_n + \text{exp}\left(\theta_k[\mathbf{e}_1]^\wedge\right)\right)\mathbf{e}_3 \nonumber 
\end{eqnarray}
with $^{k+1}\mathbf{R}_n\in SO(3)$ rotation matrix mapping the base frame $n$ into the $k+1$-th reference frame. 

The action of the stabilizer to keep the wire straight during shaping also implies that the base rotation of the guidewire $\phi$ can be assumed to seamlessly apply as a rotation around $\mathbf{z}_{k+1}$, given the natural torquability of the wire. Therefore, when the robot applies the $k$-th rotatory action $\phi_k$ to define the direction of bending for the $k$-th segment, we obtain the rotation in the base frame

\begin{eqnarray}
    ^{n}\mathbf{p}_k &=& l(n-k)\text{exp}(\phi_k[\mathbf{e}_3]^\wedge) \:^{k+1}\mathbf{R}_n \nonumber \\
    &+& l\text{exp}(\phi_k[\mathbf{e}_3]^\wedge)\text{exp}\left(\theta_k[\mathbf{e}_1]^\wedge\right)\mathbf{e}_3 
\end{eqnarray}

One can see that, by definition $^{k+1}\mathbf{R}_n = \phi_k[\mathbf{e}_3]^\wedge$, since no other rotations are allowed between the base frame and frame $k+1$; thus, we obtain the local shape of the wire from the robot's actions

\begin{equation}
    ^{n}\mathbf{p}_k  = l\left((n-k) \mathbf{I} + \text{exp}(\phi_k[\mathbf{e}_3]^\wedge)\text{exp}\left(\theta_k[\mathbf{e}_1]^\wedge\right)\right)\mathbf{e}_3 
\end{equation}
with $\mathbf{I}\in\mathbb{R}^{3\times3}$ identity matrix.

The proposed model associates the local actions of the robot with the local shape $\mathbf{p}_k$ of the $k$-th segment and can be used to convert the overall desired shape of the wire $\mathbf{p}(s)$ into the actual shape executed by the robot. However, given the complexity of medical guidewires, the exact amount of bending $\theta_k$ is hard to characterize solely using physical knowledge. Therefore, we use a data-driven approach that can be generalized to any wire, as discuss in the next section.

\section{Model Tuning}
\label{sec:model_tuning}
\label{section:tuning}

We deploy an empirical calibration protocol to identify $\theta_k$ for a given guidewire.
We apply a constant pinching force from the grippers of the robot, as indicated by the motor action $\beta_k$, and we keep the bending plane rotation constant at $\phi_k = 0, \forall k=1,2,\dots,n$. The robot is commanded to sequentially actuate $n$ consecutive segments, each of equal length, via the action $\delta_k = l, \forall k=1,2,\dots,n $. We observe a planar, discrete, constant-curvature arc at the guidewire's distal tip.

Under this protocol, each actuated segment exhibits the same unknown local bend $\theta = \theta_i=\theta_j, \forall i,j = 1,2,\dots,n$, so the chain is a sequence of identical arcs, whose net effect is a larger circular arc.
We denote the end-to-end chord of the shaped tip with \(C\) which, for a chain of \(n\) segments of length \(l\) and bending angle \(\theta\) reads as
\begin{equation}
  C \;=\; l\,\frac{\sin\!\big(\tfrac{n\theta}{2}\big)}{\sin\!\big(\tfrac{\theta}{2}\big)} .
  \label{eq:chord}
\end{equation}

For each wire we aim to shape, we can perform this process on the physical system and measure the actual cord length $\bar{C}$ experimentally. From this, the characteristic bending angle of the wire $\theta^*$ is found as the solution of the equation $C- \bar{C} = $, i.e.  

\begin{equation}
\label{eq:solve}
  l\,\frac{\sin\!\big(\tfrac{n\theta^*}{2}\big)}{\sin\!\big(\tfrac{\theta^*}{2}\big)} \;-\; \bar{C} \;=\; 0,
\end{equation}
which can be solved for $0<\theta^*<\frac{2\pi}{n}$ and $0<\bar{C}\leq nl$. This implies that appropriate tuning is obtained using shapes that deviate from straight or perfect circles.

\section{Experimental Validation}
\label{sec:experiments}
In the following section, we present the experimental evaluation of the platform using the commercially available ARROW\textsuperscript{\texttrademark} Marked Spring guidewire (\SI{0.64}{\milli\meter} diameter, shapeable tip \SI{20}{\milli\meter}, total length \SI{680}{\milli\meter}). We first perform a wire-specific calibration in Section \ref{sec:model_calibration} and consequently validate the shaping accuracy in Section \ref{sec:model_val}.

\myBlue{The model serves primarily as a kinematic framework that maps robot actions to wire geometry. Because guidewire mechanics are complex and nonlinear, characterizing the exact bend angle $\theta_k$ purely from physical parameters (e.g., Young’s modulus, friction) is unreliable. Therefore, the model is designed to be calibrated experimentally. For a new wire type, the user performs a single standardized test (shaping a constant-curvature arc) to identify the characteristic bend parameter $\theta^*$. Once this single parameter is identified, the model uses it to accurately predict complex, multi-planar geometries (e.g., S-shape, Hook, Helix) without requiring shape-specific retraining.}

\subsection{Calibration}
\label{sec:model_calibration}

\begin{figure}[t]
    \centering
    \includegraphics[scale = 0.97]{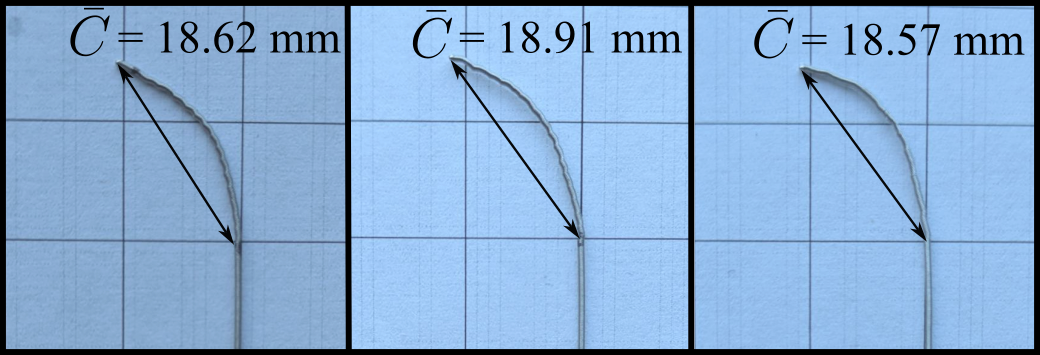}
    \caption{Measured end-to-end chord length ($\bar{C}$) of the 10 segment arc of three identical wires.}
    \label{fig:calib}
\end{figure}

The per-segment length was set to $l = 2.0$ mm, and the bending plane was held constant ($\phi = 0^\circ$). The actuator's servo-controlled pinching force was set to a pre-determined value, optimized to induce a consistent bend in the guidewire tip without causing plastic deformation or material fatigue. The robot was then commanded to sequentially actuate $n = 10$ consecutive segments. To assess the repeatability and consistency of the calibration, this entire procedure was repeated three times, using a fresh, identical ARROW guidewire for each trial. After each trial, the end-to-end chord length of the resulting 10-segment arc was measured, as illustrated in Fig. \ref{fig:calib}.

The three trials yielded a mean chord length of $\bar{C} = 18.7 \pm 0.21 \, \text{mm}$ . We then recovered the per-segment bend angle $\theta$ by substituting the known parameters ($l = 2.0$ mm, $n = 10$) and the mean measured chord length into (\ref{eq:solve}). We solved (\ref{eq:solve}) numerically using Matlab's \texttt{fsolve} function and identified a unique per-segment bend angle of $\theta \approx 6.89$\SI{}{\degree}.

We use this calibrated value in further shaping experiments described below.

\subsection{Model Validation}
\label{sec:model_val}

To validate the accuracy of our model from Section \ref{sec:wire_model}, we used the empirically-derived per-segment bend angle ($\theta \approx 6.89^\circ$) as a foundational parameter to generate and test four distinct, clinically-relevant tip shapes. 

\myRed{The 'C' shape was a planar, constant-curvature arc created by actuating $n$ consecutive segments in a constant plane ($\phi_i = 0^\circ$), similar to the one used for calibration. The 'S' shape was a planar, alternating-curvature geometry, formed by actuating the proximal segments at $\phi_i = 0^\circ$ and flipping the bending plane midway to $\phi_i = 180^\circ$ for the distal segments ($i > n/2$). The ``Angled" shape was created by actuating only the distal-most $k < n$ segments in a single plane, leaving the proximal portion straight. Finally, the ``Hook" shape was geometry formed by leaving an initial proximal set of segments unactuated, bending a central set at $\phi_i = 0^\circ$, and then bending the final distal set in the opposite direction at $\phi_i = 180^\circ$.}

\myBlue{The 'C' shape was a planar, constant-curvature arc created by actuating $n$ consecutive segments in a constant plane ($\phi_i = 0^\circ$), similar to the one used for calibration. This fundamental geometry was selected as it represents the standard 'primary curve' used in most clinical guidewires for general navigation and simple vessel selection. The 'S' shape was a planar, alternating-curvature geometry, formed by actuating the distal segments at $\phi_i = 0^\circ$ and flipping the bending plane midway to $\phi_i = 180^\circ$ for the proximal segments ($i > n/2$). This profile was chosen to test the system's ability to transition between opposing bending planes, mimicking the double-curves often required to traverse sigmoid anatomies such as the carotid siphon. The ``Angled" shape was created by actuating only the distal-most $k < n$ segments in a single plane, leaving the proximal portion straight. Finally, the ``Hook" shape was geometry formed by leaving an initial distal set of segments unactuated, bending a central set at $\phi_i = 0^\circ$, and then bending the proximal set in the opposite direction at $\phi_i = 180^\circ$. This geometry was intended to replicate 'recurved' or 'shepherd’s crook' shapes, which are widely considered valuable for cannulating vessels with acute takeoff angles that are difficult to access with simple arcs.}

\begin{figure}[t]
    \centering
    \includegraphics[scale=0.97]{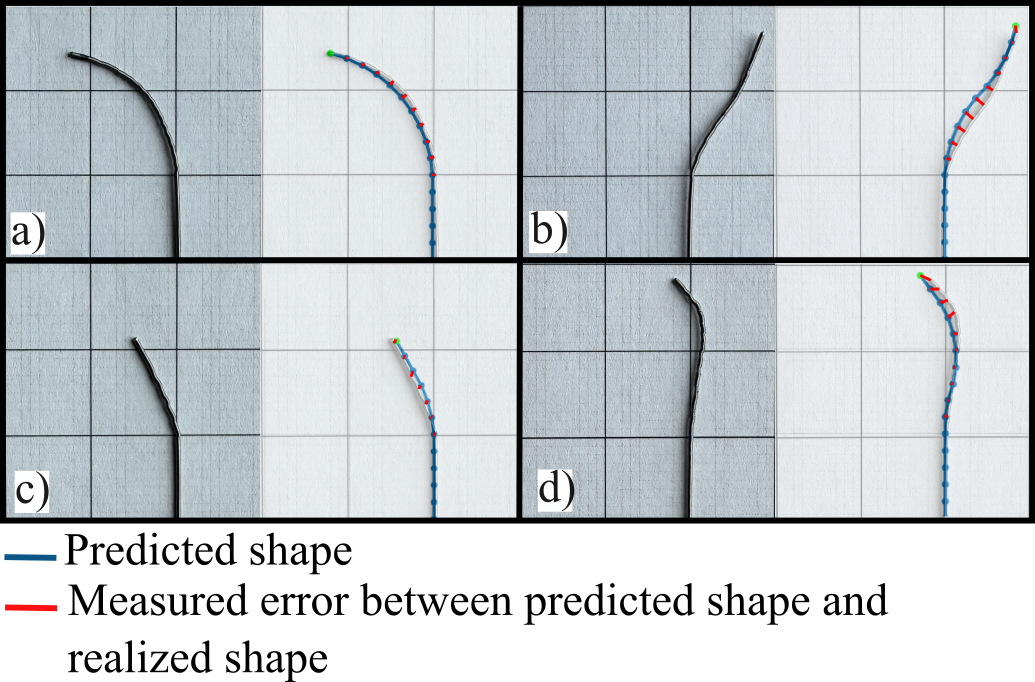}
    \caption{Predicted (model) versus realized centerlines for the a) \emph{C}, b) \emph{S}, c) \emph{Angled}, and d) \emph{Hook} shapes using the calibrated bend \(\hat{\theta}\).}
    \label{fig:shapes}
\end{figure}

For each shape, a sequence of commands was sent to the robotic platform, and the resulting physical guidewire shape was measured. This measured shape was then compared to the shape predicted by our model using the identical command sequence.
Figure~\ref{fig:shapes} shows the images used to quantify the difference between predicted and measured shapes. In each of the figures, the right-most depiction of each shape shows the model prediction (blue) \myRed{overlapped with the measured shape in red} \myBlue{overlaid on the measured shape, with the error depicted in red} \myReview{R2-1}. We can qualitatively notice that the desired shapes are, in fact, replicated by the proposed model. 

\begin{figure}[t]
    \centering
    \includegraphics[width=\columnwidth]{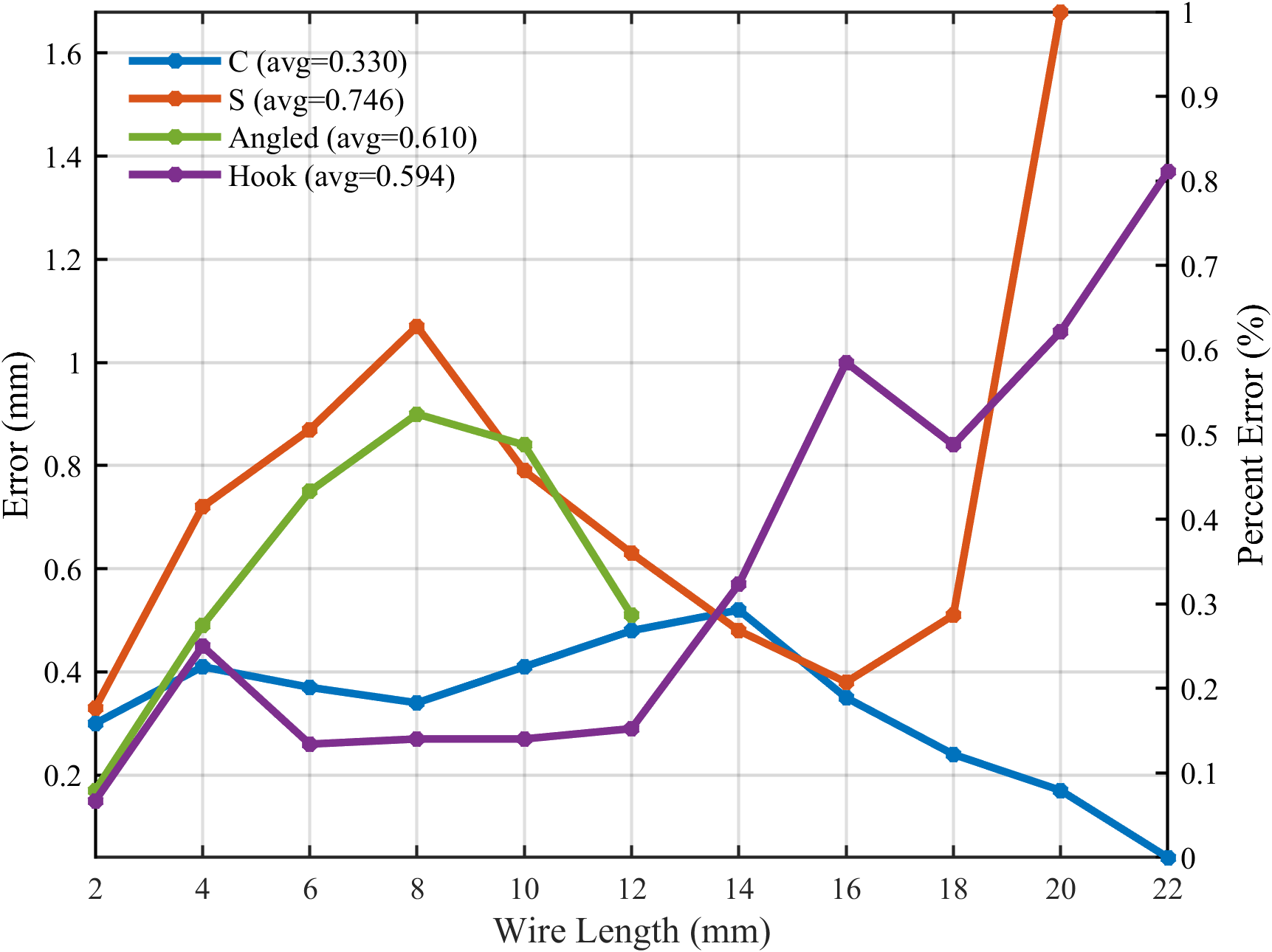}
    \caption{Segment-wise errors across wire shapes.}
    \label{fig:errorplot}
\end{figure}

For quantitative analysis, the predicted and measured point sets, $\mathbf{p}^{\text{pred}}$ and $\mathbf{p}^{\text{meas}}$ respectively, were aligned to a common base frame. We then computed the per-segment planar Euclidean error $e_k=\big\|\mathbf{p}^{\text{meas}}_k-\mathbf{p}^{\text{pred}}_k\big\|_2$ for each segment $k=1,\dots,n$. The $\min$–$\max$ and mean errors for each of the four shapes are reported in Table~\ref{tab:error_per_shape}. Figure~\ref{fig:errorplot} plots $e_k$ versus the wire length in both \SI{}{mm} and \% with respect to the total shapable length of the wire.

\myBlue{While the mean error remains low (\SI{0.56}{\milli\meter}), the larger errors are characteristic of open-loop dead-reckoning, where small kinematic uncertainties at the base propagate to the tip.} Among the tested geometries, the planar 'C' shape, which requires no inter-segmental changes in the bending plane, yielded the lowest errors. In contrast, shapes requiring out-of-plane phase changes or alternating bend directions ('S', 'Angled', and 'Hook') incurred higher distal errors. This is expected, as the final accuracy of these complex shapes depends critically on both the bend magnitude $\theta$ and the precision of the axial phase $\phi_i$ applied at each segment.

\myBlue{These results demonstrate that the mechanism is fundamentally capable of the necessary shaping actions, while the specific outliers highlight the necessity of the closed-loop compensation strategies proposed for future work to achieve clinical-grade precision.} Overall, the model consistently captures the centerline geometry to sub-millimeter accuracy near the base and maintains millimeter-scale fidelity distally.

\begin{table}[t!] 
    \vspace*{15pt}
  \centering
  \caption{Errors between measured and modeled shapes extracted using visual analysis of the pictures in Fig. \ref{fig:shapes}}
  \label{tab:error_per_shape}
  \resizebox{\columnwidth}{!}{%
  \begin{tabular}{l c c}
    \toprule
    \textbf{Shape} & \textbf{Error (min--max) [mm]} & \textbf{Average [mm]} \\
    \midrule
    C           & 0.04--0.52 & 0.33 \\
    S           & 0.33--1.68 & 0.70 \\ 
    Angled      & 0.17--0.90 & 0.61 \\ 
    Hook        & 0.15--1.37 & 0.59 \\
    \midrule
    \textbf{Average} & -- & 0.56 \\ 
    \bottomrule
  \end{tabular}%
  }
\end{table}

\subsection{Demonstration of Navigation In-vitro}
\label{sec:anatomy}

\begin{figure}
    \centering
    \includegraphics[width=\columnwidth]{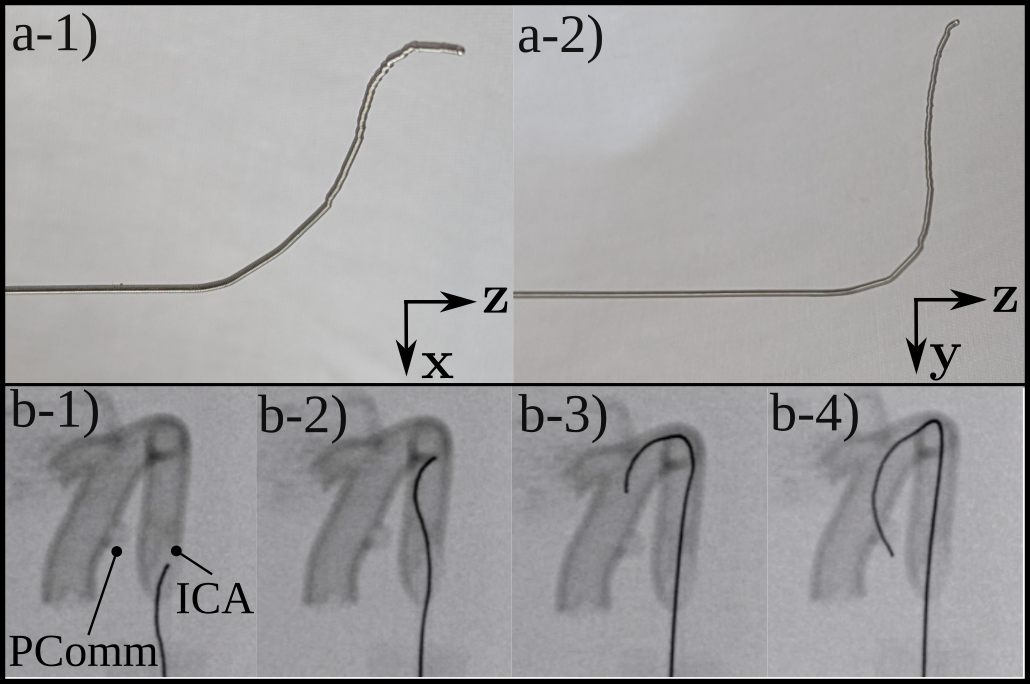}
    \caption{Demonstration of navigation from the \gls{ica} to \gls{pcom}.}
    \label{fig:3Dshapes}
\end{figure}

Beyond the planar tips in Section~\ref{sec:model_val}, we evaluated a multi-planar helical tip to assess whether 3D shaping can bias passive navigation at cerebrovascular bifurcations. The helix was generated by repeating a roll–bend–advance primitive with constant per-segment axial rotation \(\Delta\phi_k = \phi_k - \phi_{k-1}\) and the calibrated local bend \(\theta_k\) over a segment length \(l=\SI{2}{\milli\meter}\). In this study, we used a shallow helix with \(\Delta\phi_k=\ang{45}\) per segment, as shown in Fig.~\ref{fig:3Dshapes}a. This helical shape produces a large effective radius of curvature while preserving distal steerability. 

Tests were conducted on a portion of the Circle of Willis 3D printed in clear material using an SLA printer (Form 4, Formlabs). The evaluation segment comprised the right \gls{ica}, including the carotid siphon and the distal decision point, where one branch proceeds into the \gls{mca}, while the other follows the \gls{pcom} route toward the \gls{pca}. We demonstrate navigation using the helically-shaped wire to avoid entering the \gls{mca} and to enter the \gls{pcom}. This navigation is particularly complicated, as depicted in Fig. \ref{fig:main_figure}, due to the several turns that the wire needs to navigate. Its accomplishment is also fundamental from a clinical standpoint, since reaching the \gls{pcom} may be required in several procedures, such as aneurysm embolization \cite{Ahmed2014NavigationEmbolization} and mechanical thrombectomy \cite{Otsuji2017PureThrombectomy}. 

The guidewire was positioned at the proximal \gls{ica} origin and advanced manually with two degrees of freedom—axial push and shaft rotation—while allowing the wire to follow the vessel passively. Trajectories were recorded under fluoroscopy for subsequent analysis, as shown in the Supplementary Video. Figure~\ref{fig:3Dshapes} reports 4 significant snapshots: (i) start configuration; (ii) wire approaching the first bend past the Cavernous into the Supraclinoid segments of the \gls{ica}; (iii) wire exiting the first bend and avoiding the \gls{mca}; (iv) wire entering the \gls{pcom}.

In contrast to standard primary curves, which typically overshoot the acute takeoff of the \gls{pcom} and track into the larger \gls{mca}, we observed that, the helical tip  glided along the curvature of the carotid siphon and naturally continued into the \gls{pcom} toward the \gls{pca}. Qualitatively, this behavior is consistent with the helix’s larger effective turning radius we designed, which matches the gradual curvature of the siphon and \gls{pcom} origin.

These results suggest that programmable 3D tip shaping  can bias passive navigation through complex geometries (siphons, acute bifurcations) without increasing the operator's burden. Practically, a shallow helix offers a distal shape that promotes smooth tracking through gently curving sections and preferential alignment with the \gls{pcom} pathway, potentially reducing trial-and-error during distal access.

\section{Conclusions}
\label{sec:conclusions}


We presented a robotic guidewire shaping system designed to standardize and automate the process of neuroendovascular wire shaping, which is traditionally performed manually and heavily relies on surgeon expertise. This system represents a step toward reducing inter-operator variability and enabling reproducible, anatomy-specific wire configurations.

We present a data-driven kinematic model that maps robot actions to the resulting wire shapes. This model not only enables forward prediction of wire geometries and also lays the groundwork for future inversion to allow computer-generated or image-derived shapes to be translated into actionable robot commands. Our experimental validation demonstrated that the model achieves a \gls{rms} error of 0.56 mm across a range of clinically relevant 2D tip shapes, including C, S, Angled, and Hook configurations.

We further demonstrated the system’s capability to perform 3D shaping and validated its functionality through an in-vitro navigation task. Specifically, we shaped a wire to traverse a complex anatomical path from the petrous segment of the \gls{ica} to the \gls{pcom}, showcasing the system’s potential for real-world neurovascular applications.

Future work will focus on integrating patient-specific, image-based shape computation to enable personalized wire shaping based on our previous work \cite{Tamhankar2025TowardsPlanning}. We will also develop an inverse kinematic model to directly map desired shapes into robot actions and improve the shaping precision and repeatability of the system. To actively compensate for the geometric deviations observed in Section \ref{sec:model_val}, we intend to transition to a closed-loop control framework. By integrating real-time feedback, the system will perform online calibration to adapt to material variability and apply active error correction during the shaping process.
Further performance evaluation will also be performed across different wire types, and eventually, we will conduct in-vivo studies to assess clinical feasibility and impact.


\bibliographystyle{IEEEtran}
\bibliography{references,references2}											
\end{document}